\documentclass{article}

     \PassOptionsToPackage{authoryear,round}{natbib}
\usepackage[final]{neurips_2024_ml4ps}

\usepackage[utf8]{inputenc} % allow utf-8 input
\usepackage[T1]{fontenc}    % use 8-bit T1 fonts
\usepackage[colorlinks]{hyperref}       % hyperlinks hide links
\hypersetup{
 linkcolor=sthlmRed
,citecolor=sthlmGreen
,urlcolor=sthlmBlue
}
\usepackage{url}            % simple URL typesetting
\usepackage{booktabs}       % professional-quality tables
\usepackage{amsfonts}       % blackboard math symbols
\usepackage{nicefrac}       % compact symbols for 1/2, etc.
\usepackage{microtype}      % microtypography
\usepackage{xcolor}         % colors
\usepackage{amsmath}
\usepackage{amsfonts}
\usepackage{mathrsfs}
\usepackage{amssymb}
\usepackage{tabularray}
\usepackage{amsthm}
\usepackage{mathabx}
\usepackage{array,multirow,makecell}
\usepackage[mathcal]{euscript}
\usepackage[normalem]{ulem}
\usepackage{stmaryrd}
\usepackage{bbold}
\usepackage{paralist}
\usepackage{todonotes}

\usepackage[capitalize,noabbrev]{cleveref}
\usepackage{enumerate}
\usepackage{algorithm}
\usepackage{algorithmic}
\usepackage{wrapfig}
\usepackage{subfig}

\usepackage{thmtools}
\usepackage{thm-restate}
\usepackage{hyperref}
\usepackage{cleveref}

\usepackage{pgfplots}
\usepackage{tikz}
\usetikzlibrary{shapes}
\usetikzlibrary{snakes}
\usetikzlibrary{arrows}
\usetikzlibrary{automata}
\usetikzlibrary{fit}
\usetikzlibrary{shadows}
\usetikzlibrary{trees}
\usetikzlibrary{decorations}
\usetikzlibrary{decorations.text}
\usetikzlibrary{positioning}
\usetikzlibrary{patterns}
\usetikzlibrary{backgrounds}
\usetikzlibrary{matrix}
\usetikzlibrary{arrows.meta}
\usetikzlibrary{calc, shapes}
\usetikzlibrary {petri} 
\usetikzlibrary{chains,scopes}

\usepackage{xspace}

	\definecolor{sthlmLightBlue}{RGB}{214,237,252} % HEX #d6edfc
		
	\definecolor{sthlmBlue}{RGB}{0,110,191} % HEX #006ebf

	\definecolor{sthlmLightGreen}{RGB}{213,247,244} % HEX #0d5f7f4
		
	\definecolor{sthlmGreen}{RGB}{0,134,127} % #00867f

	\definecolor{sthlmLightGrey}{RGB}{213,217,225} %HEX #D5D9E1
		
	\definecolor{sthlmGrey}{RGB}{245,243,238} % HEX #f5f3ee
		
	\definecolor{sthlmDarkGrey}{RGB}{51,51,51} % HEX #333333

	\definecolor{sthlmLightOrange}{RGB}{255,215,210} % HEX #ffd7d2
		
	\definecolor{sthlmOrange}{RGB}{221,74,44} % HEX #dd4a2c

	\definecolor{sthlmLightPurple}{RGB}{241,230,252} % HEX #f1e6fc
		
	\definecolor{sthlmPurple}{RGB}{93,35,125} % HEX #5d237d

	\definecolor{sthlmLightRed}{RGB}{254,222,237} % HEX #c40064
		
	\definecolor{sthlmRed}{RGB}{196,0,100} % HEX #fedeed

	\definecolor{sthlmYellow}{RGB}{252,191,10} % HEX #fcbf0a

\usepackage{colortbl}%
  
\usepackage{booktabs}

\title{Video-Driven Graph Network-Based Simulators}

\author{%
  Franciszek Szewczyk \\
  University of Groningen \\
  \texttt{f.szewczyk@student.rug.nl}\\
   \And
   Gilles Louppe \\
   University of Liège \\
   \texttt{g.louppe@uliege.be} \\
   \And
   Matthia Sabatelli \\
   University of Groningen \\
   \texttt{m.sabatelli@rug.nl} \\
}

\begin{document}

\maketitle

\begin{abstract}
Lifelike visualizations in design, cinematography, and gaming rely on precise physics simulations, typically requiring extensive computational resources and detailed physical input. This paper presents a method that can infer a system's physical properties from a short video, eliminating the need for explicit parameter input, provided it is close to the training condition. The learned representation is then used within a Graph Network-based Simulator to emulate the trajectories of physical systems. We demonstrate that the video-derived encodings effectively capture the physical properties of the system and showcase a linear dependence between some of the encodings and the system's motion.
\end{abstract}

\section{Introduction}

Realistic simulations of physical processes are crucial in engineering for validating product integrity and functionality under extreme conditions. Major animation studios rely on physics simulations to craft internationally acclaimed films \citep{mpmDisney}. Similarly, game developers aim for immersive experiences with realistic physics. Each one of these applications imposes distinct requirements: game designers prioritize real-time performance, while filmmakers and engineers prefer higher accuracy at the cost of longer computation times. While advances in computational power have enabled more complex simulations, traditional methods require detailed physical input and domain expertise. In this work, we aim to overcome these limitations by creating a system capable of simulating various materials based solely on short video clips. Once trained, our model can be presented with a short video of a physical system to infer its physical properties, which are later used in a Graph Network-Based Simulator \citep{learningToSimulateComplexPhysics} framework to predict the motion of various systems. We find that the encodings generated from the videos differentiate between different physical properties of the recorded system and that there is a linear correspondence between the physical encoding and the final prediction of the model.

%=====================================================

\paragraph{Preliminaries}
\label{sec:preliminaries}

Employing notations akin to \citet{relationalInductiveBiasesGraphNetworks} we define a graph as a tuple $G=(V,E)$, where $V=\{v_i\}_{i=1:N^v}$ is the set of vertices, with $v_i$ denoting the attributes of the vertex and $E=\{(e_k,r_k,s_k)\}_{k=1:N^k}$ is the set of edges with $e_k$ being the edge's attribute, $r_k$ the index of the receiver node, and $s_k$ the index of the sender node. Each edge encodes the direction and the distance between two vertices. A GNS, modeled by a graph neural network, denoted as $g_\phi : (\mathcal{X}, \mathcal{P}) \rightarrow \mathcal{Y}$, maps a state of the system $X^t \in \mathcal{X}$ to per-particle accelerations - $Y^t \in \mathcal{Y}$. Each vertex attribute $v_i$ is a concatenation of two components: the velocity of the particle $i$ in each of the last $C \in \mathbb{N}^+$ simulation steps and a physical encoding $P \in \mathcal{P}$, which represents the physical properties of the material, allowing the network to infer various types of systems.
The subsequent state $X^{t+1}$ is then derived using a semi-implicit Euler integrator. The processing within the GNS typically encompasses three key stages: i) encoding, ii) processing, and iii) decoding. The encoding stage maps the original graph into latent space.
The processing stage consists of $M$ message passing steps.
Finally, in the decoding phase, the final vertex attributes are decoded to obtain per-particle accelerations. The Graph Network is trained on a dataset compiled from multiple trajectories of a physical system. While \citet{learningToSimulateComplexPhysics} explicitly provide the system's physical encoding, $P$, to the model, it is worth noting that this information may not always be available. We will now propose an extension to the GNS framework that involves implicitly learning this physical encoding thanks to a short video of the system.

\section{Video-Driven GNS}
\label{sec:vegn}

\begin{wrapfigure}{r}{0.4\textwidth}
    \centering
    \includegraphics[width=\linewidth]{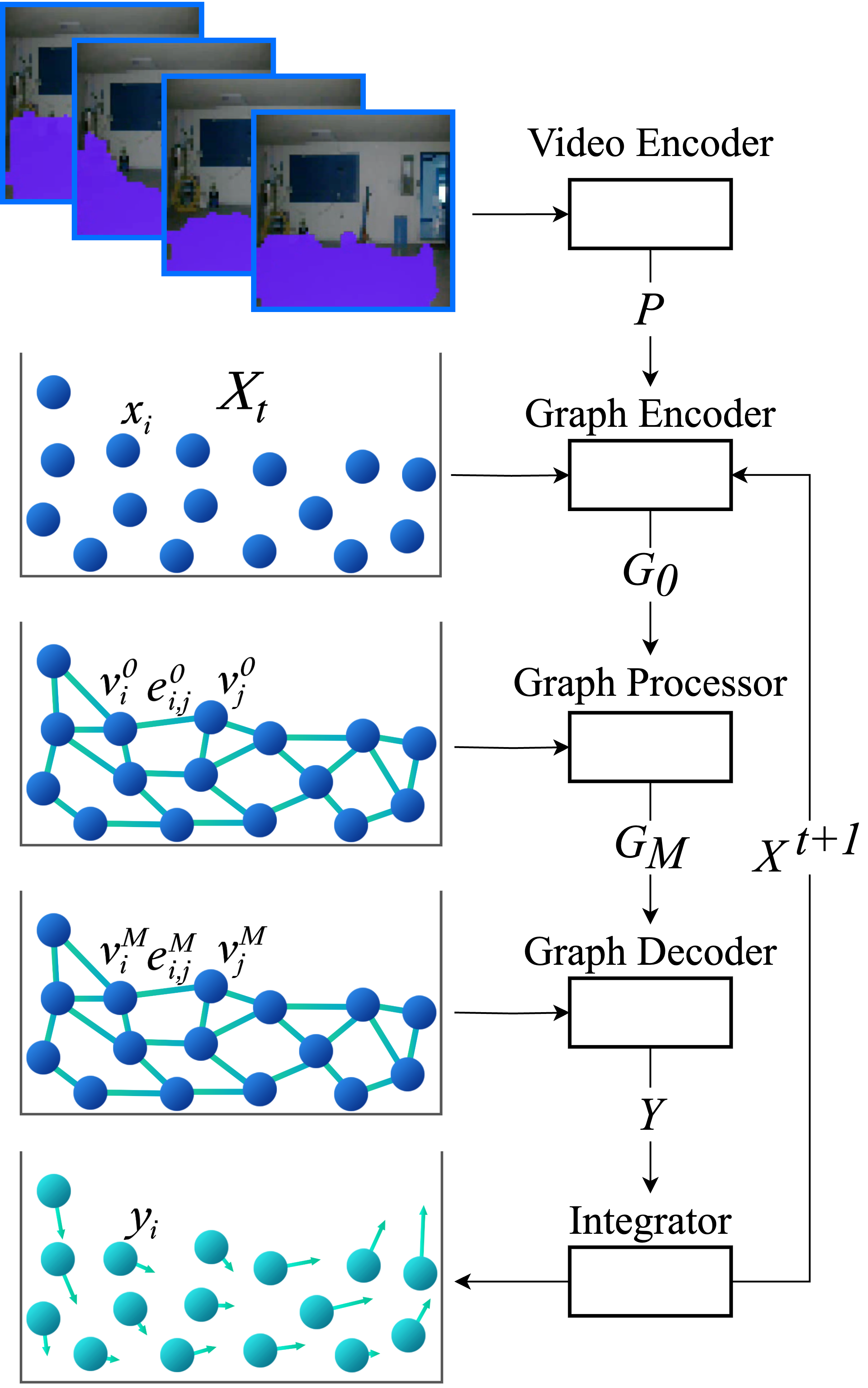}
    \caption{The video encoding $P$ is combined with vertex attributes to form the initial graph $G_0$. Message passing, performed by the Graph Processor, iterates $M$ times. Subsequently, dynamic data is decoded, and particle positions are updated using an integrator.}
    \label{fig:general-gn-diagram}
\end{wrapfigure}
Our goal is to construct an architecture that is capable of inferring the motion of diverse physical systems from short videos: we call this modification of the original GNS the Video-Driven Graph Network-based Simulator (VDGNS). Our methodology comprises two principal components: a \textsc{Video Encoder} responsible for deducing the physical encoding, denoted as $P$, and a regular GNS as described above. The architectural diagram is depicted in Figure \ref{fig:general-gn-diagram}. Our approach is remarkably straightforward: given a sequence of $n$ frames representing the visual representation of system $S$ as $F^{t_1:t_n}=(F^{t_1},...,F^{t_n}) \in \mathcal{F}$, we define a \textsc{Video Encoder}, denoted as $f : \mathcal{F} \rightarrow \mathcal{P}$, where $P \in \mathcal{P}$ represents a vector containing the latent representation of the physical properties of $S$. Once the models are trained, this formulation allows us to present the system with a short video input and infer a full trajectory of a system given some initial condition. The entire system can be trained jointly by regressing over the final predicted accelerations using standard gradient-based optimization. It is important to note that during training, we ensure that the output of the \textsc{Video Encoder} solely depends on the physical attributes of the system rather than the particular motion type or direction depicted in the video. To achieve this, we employ a customized batching procedure. Initially, we sample a state $X^t$ and its corresponding target accelerations $Y^t$ from a trajectory of some system. Then, we randomly select one of the videos depicting that particular system class, denoted as $F^{t_1:t_n}$, irrespective of whether  $X^t$ and $Y^t$ sampled earlier are part of the same trajectory. This approach ensures that the only linkage between the system's state and the video lies in the dynamical properties of that specific system, facilitating the \textsc{Video Encoder} in learning appropriate embeddings. Generally, one is free to choose a specific implementation of the functions inside of the GNS as well as the architecture of the \textsc{Video Encoder}. In our experiments, we follow the original approach \citep{learningToSimulateComplexPhysics} and implement all encoding, processing, and decoding functions of the GNS as Multi-Layer Perceptrons (MLPs). In the \textsc{Video Encoder}, the frames are encoded into a low-dimensional representation before being passed into a Long Short-Term Memory network (LSTM) (\citealp{lstmOriginalPaper}), an approach based on \citet{learningBasedClothMaterialRecovery}.

\section{Experiments}

\paragraph{Benchmark}  We compare our architecture to a baseline model that directly uses a one-hot encoding of the system class, bypassing the approximation of the physical encoding $P$. This allows us to treat the baseline as the performance limit for our architecture. The model's performance is evaluated on a task involving granular materials comprising hundreds of particles within a two-dimensional space. Rather than assessing the interpolation capabilities of the \textsc{Video Encoder}, we focus on the GNS by treating it as a trajectory generator for diverse systems, including those unencountered during training. 

\paragraph{Model}
The \textsc{Video Encoder} processes flattened video frames using a 3-layer MLP with \texttt{tanh} activations, producing a 192-dimensional encoding for each frame. This sequence of encodings is then passed through a single LSTM block, which outputs a 4-dimensional vector representing the physical encoding $P$. The dimensionality of $P$ is chosen to match the one-hot encoding of the system class for a fair comparison against a baseline model. The GNS operates on a latent graph representation, where node and edge attributes are 48-dimensional vectors. The graph processing stage includes $M=3$ message-passing steps. A detailed account of our architecture is provided in Appendix \ref{ap:model}.

%Specifically, we train the models on four distinct classes of systems: \texttt{water}, \texttt{sand}, \texttt{snow}, and \texttt{elastic}. For the exact physical parameters, we refer the reader to \citet{taichiMPM}. These system classes inherently lack interpolability. For instance, \texttt{snow} particles form clumps, a characteristic absent in other systems. Rather than quantitatively assessing the accuracy for systems not included in the training dataset, we focus on the representations generated by the \textsc{Video Encoder}. Subsequently, we sample points within this latent space, feed them into the GNS and attempt to reconstruct the physical properties of the interpolated system.

\paragraph{Task \& Datasets}

We consider four distinct classes of systems: \texttt{water}, \texttt{sand}, \texttt{snow}, and \texttt{elastic}, simulated using the Taichi-MPM simulator \cite{taichiMPM}. Most of these system classes inherently lack interoperability. For instance, \texttt{snow} particles form clumps, a characteristic absent in other systems. The \texttt{elastic} particles rigidly maintain their internal structure, resulting in a stiff object that rebounds from the environment's boundaries, behaving similarly to a rubber ball. Rather than quantitatively assessing the accuracy for systems not included in the training dataset, we focus on the representations generated by the \textsc{Video Encoder}. Subsequently, we sample points within this latent space, feed them into the GNS and analyze the impact of the video encoding on final predictions.

For training, we generate 30 trajectories for each of the four aforementioned classes. Each trajectory comprises 400 steps, covering 4 seconds in the simulation. At the onset of each trajectory, a circle of particles with a random radius is positioned randomly within the simulation boundaries. Due to computational constraints and the highly complex behavior observed at the initial stages of the simulation, the first second is excluded from both training and evaluation datasets. Drawing inspiration from \citet{learningBasedClothMaterialRecovery}, the system's rendering is overlayed on one of the 10 images selected from the dataset of indoor scene images \cite{recognizingIndoorScenes}. Limiting the number of images ensures consistent backgrounds across classes, mitigating the risk of the \textsc{Video Encoder} learning the backgrounds rather than the actual physical systems. To introduce complexity, the transparency of the overlay varies randomly between 50\% and 100\% across trajectories. The rendering color of the physical material is chosen randomly to increase dataset diversity. The videos are rendered at 20 frames per second with a resolution of $64\times64$ in the RGB color space. Pixel values are normalized between 0 and 1. The generated graphs contain the $C=3$ most recent velocities and edges are formed between two particles whenever the distance between them is smaller than 12\% of the simulation width. During each batch generation, 8 trajectory steps are sampled. Each of these steps is randomly paired with one of the videos containing the same system class. We note that this procedure creates no linkage between a trajectory step and the video of the same trajectory. For robustness, a normally distributed noise with a standard deviation of 0.05 is added to the edge embeddings. Similarly, the velocities included in the node embeddings are augmented with a similar noise with a standard deviation of 0.002. These values were chosen to alter the embeddings by roughly 20\%. For the evaluation, we generate 30 more trajectories using a similar procedure, but without adding artificial noise.

\paragraph{Metrics}
\label{sec:experiments-metrics}
We quantitatively evaluate the performance of VDGNS by reporting the Acceleration Error and the Rollout Error. The first one aims to measure the system's instantaneous accuracy and calculates the Mean Squared Error (MSE) between the true and predicted acceleration vectors to measure a one-step error. The long-term accuracy assessment is performed by sampling an initial condition and a random video of the same physical system for each trajectory in the test set. Then, we infer the rest of that trajectory using the trained model. We take inspiration from \cite{learningToSimulateComplexPhysics} and their use of optimal transport (\citealp{topicsInOptimalTransportation}) to calculate the Wasserstein distance using the Euclidean norm between the true and predicted trajectory states.

\section{Results}

\begin{table}
    \centering
    \caption{Performance of the models on the Fluid dataset. The results are averaged across the whole evaluation dataset.}
    \begin{tabular}{c|cc|c}
        System & \multicolumn{2}{c|}{\makecell{One-step MSE $[\times 10^{-8}]$}} & \makecell{Encoding\\variance} \\ \hline
               & VDGNS & Baseline & VDGNS  \\ \hline
        \texttt{Water}  & $11.38 \pm 0.00026$  & $8.64 \pm 0.00017$ & 0.183  \\
        \texttt{Snow}   & $4.65 \pm 0.0003$ & $3.36 \pm 0.00008$  & 0.152 \\
        \texttt{Sand}   & $6.06 \pm 0.00024$  & $4.65 \pm 0.00013$ & 0.191  \\
        \texttt{Elastic}& $5.81 \pm 0.00014$ & $5.01 \pm 0.00005$  & 0.066 \\
    \end{tabular}
    \label{tab:fluid-results}
\end{table}
 
\begin{figure}
    \centering
        \includegraphics[width=0.95\linewidth]{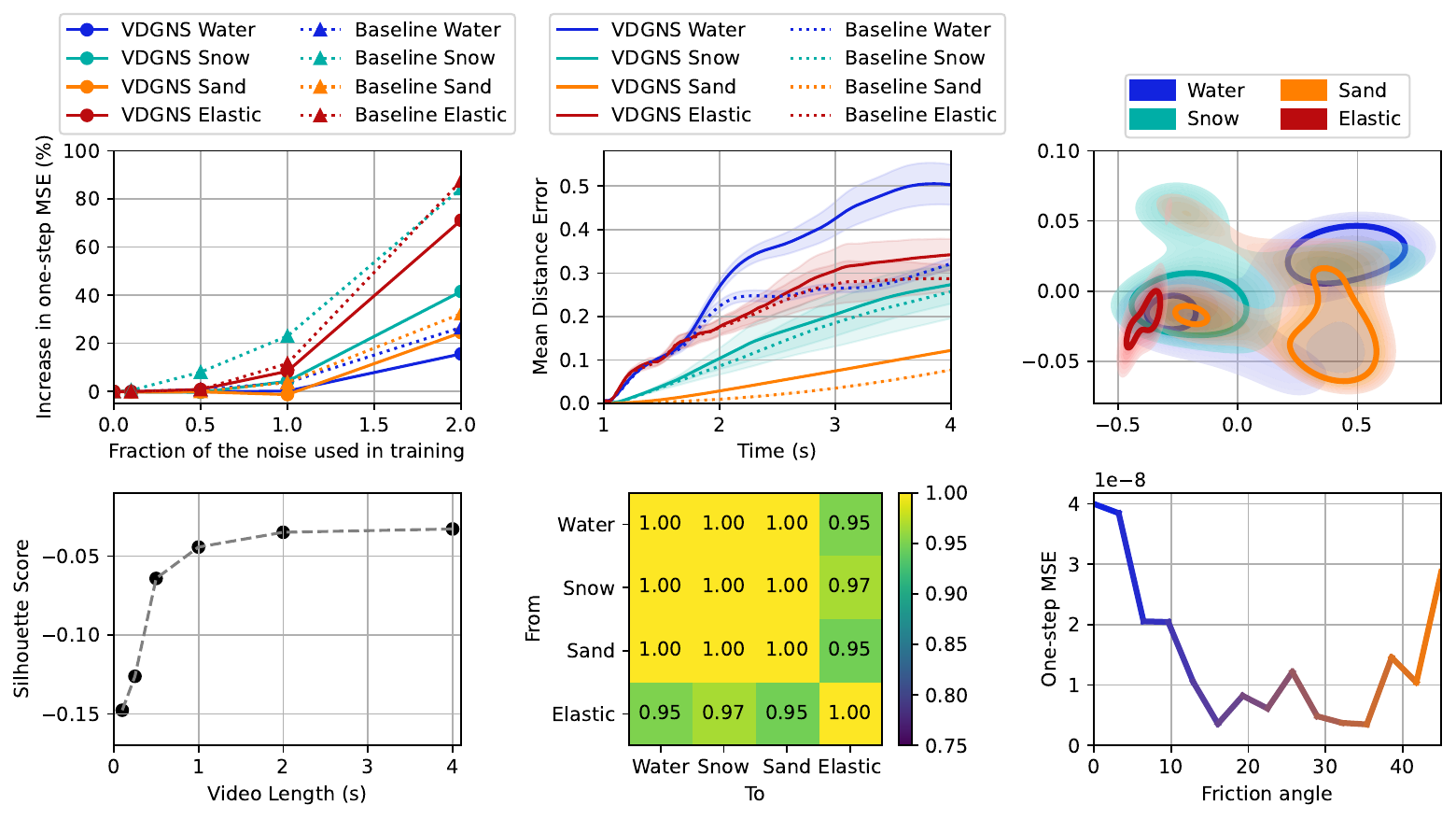}
    \caption{Top-left: Percentage increase in one-step MSE as noise level increases, relative to the MSE at zero noise. Top-middle: Mean Wasserstein distance between particles in true and predicted trajectories. Top-right: Kernel density estimate of the first two principal components of video encodings for each of the four classes. Thick lines represent the 85th percentile density contours. Bottom-left: Silhouette scores of video encodings for videos of varying lengths. Bottom-middle: The $R^2$ score of the linear relation between the video encodings and the predicted accelerations. Bottom-right: one-step MSE for varying angles of friction, showing the interpolation between water and sand.}
    \label{fig:res1}
\end{figure}

First, the one-step MSE results in Table \ref{tab:fluid-results} showcase a similar performance of VDGNS and the Baseline. As shown in the top-left plot of Figure \ref{fig:res1}, both models experience an increase in one-step MSE as noise is added to the evaluation dataset. Noise levels are reported as fractions of the noise used during training. Models' robustness differs between materials, but overall, both VDGNS and Baseline show similar trends in performance degradation as noise increases. The top-middle plot of Figure \ref{fig:res1} further shows that the Baseline performs better than the VDGNS, but both models behave similarly.

Next, we also investigate the effectiveness of the physical encodings generated by the \textsc{Video Encoder}. The top-right plot of Fig. \ref{fig:res1} shows a kernel density estimate of video encodings for each system class, reduced to 2 dimensions using Principal Component Analysis. The results reveal the model's ability to produce meaningful representations. For example, the embeddings for water and sand systems are similar due to lateral particle dispersion upon impact, while elastic and snow systems also show comparable embeddings, likely because snow, like elastic systems, can bounce upon impact before dispersing. 

The quality of video encodings is influenced by the length of the video. In the bottom-left plot of Fig. \ref{fig:res1}, we present the silhouette scores of the video encodings as a function of video length. The results show that longer videos yield better separation and higher quality in the encodings for each system class. However, the improvement in encoding quality diminishes significantly for videos longer than one second.

To analyze the impact of video encodings on the final predictions, we first compute the mean encoding for each system class. Then, we use linear interpolation to generate 10 intermediate encodings between each pair of classes. Next, we analyze how these encodings impact the predicted per-particle accelerations in the evaluation dataset. We focus on the linearity between the video encodings and the final predictions. The mean $R^2$ score is calculated by comparing predictions from the interpolated encodings to the linear interpolation of the predictions for each class. These results are shown in the bottom-middle plot of Fig. \ref{fig:res1}, indicating a strong linear relationship across all classes.

Finally, we analyze the interpolation capabilities of our model. While the system classes lack interoperability, we exploit the implementation of the simulator. The \verb|sand| class has a parameter for the angle of friction, set to 45 degrees by default. When set to 0, the system behaves like the \verb|water| class. To test the model's ability to interpolate between systems, we create an additional evaluation dataset. This dataset includes 4 trajectories for 15 system classes, with the friction angle increasing from 0 to 45 degrees. We then evaluate the system's one-step accuracy on this dataset. The bottom-right plot in Figure \ref{fig:res1} shows the one-step errors for varying angles of friction.

\section{Discussion \& Conclusion}

Our proposed VDGNS architecture performs very similarly to the baseline despite being trained without being explicitly provided with the system's physical properties. While the results obtained by the baseline are generally slightly better, it is worth noting that when investigating the mean distance errors between the particles over entire trajectories, both models exhibit very similar and low rollout error patterns. Furthermore, the models are robust to a limited amount of noise, although some materials are more sensitive than others. These findings support the notion that the \textsc{Video Encoder} is capable of extracting meaningful representations that the GNS can make use of. Importantly, the video encodings clearly distinguish between the physical properties of the system. Further, we consider the main limitation of our approach to be the need for prior knowledge of the system classes that the videos belong to. Specifically, each video and particle trajectory in the dataset must be manually assigned a class, reducing the method's generalization capabilities. Unsupervised learning techniques could be used to overcome this shortcoming. We believe that the generation of such video encodings based on real, not simulated, data will further improve the effectiveness and applicability of the GNS framework. 

%An important observation is that the mean encodings generated by the \textsc{Video Encoder} correlate with the system's dynamics, even when encountering novel physical parameters, despite not being explicitly trained on them. However, we acknowledge that the variance of these encodings is substantial when inferring previously unencountered systems. To mitigate this issue, we propose averaging the physical encodings of multiple videos of the same system, as the mean encodings exhibit a clear correlation with the true physical properties. 

%=====================================================

\bibliography{mybib}
\bibliographystyle{plainnat}

\appendix

\newpage
\section{Additional Figures}

\begin{figure}[h]
    \centering
    \includegraphics[width=0.4\linewidth]{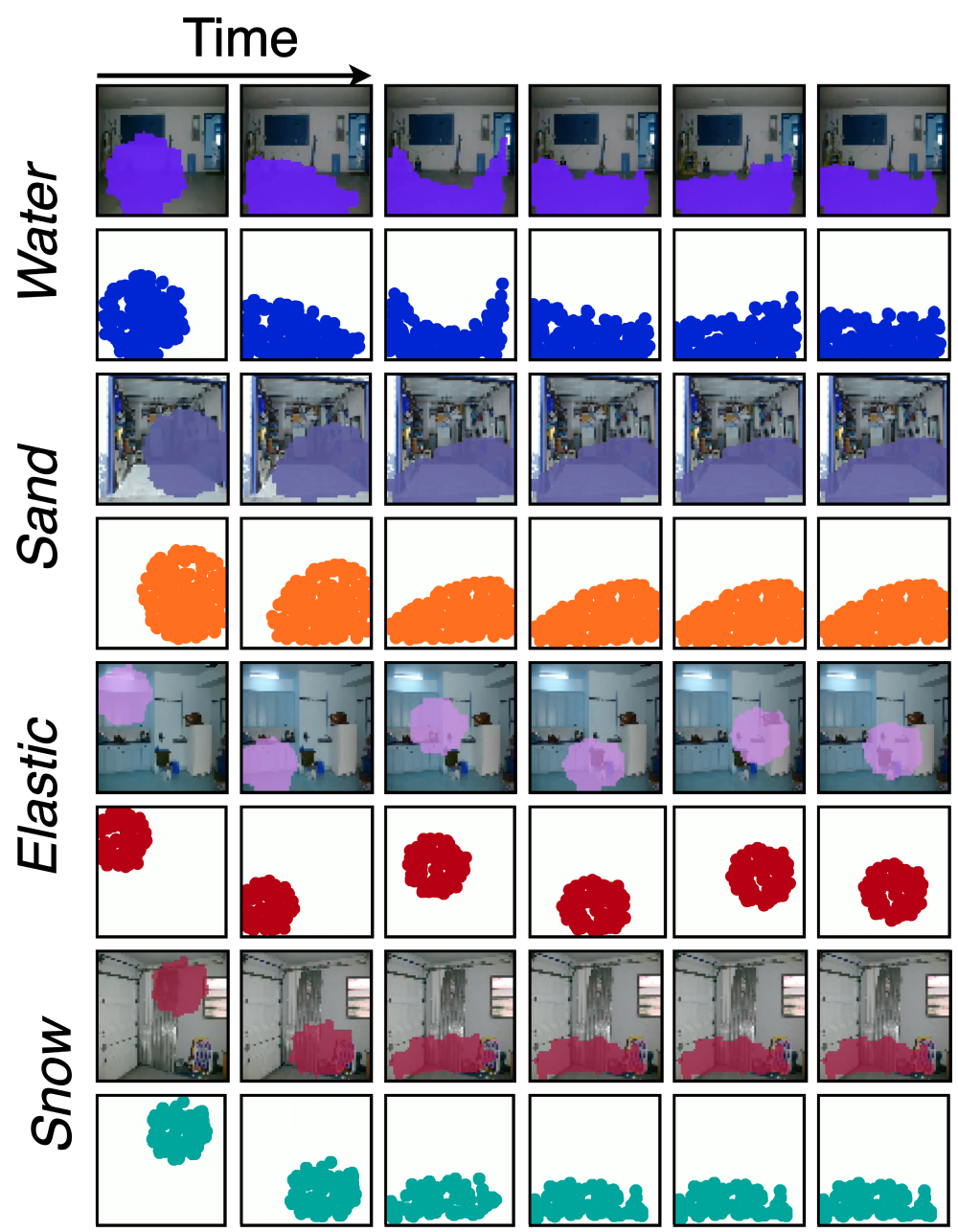}
    \caption{Sample trajectories and corresponding videos for each of the four classes. Time flows right.}
    \label{fig:dataset-trajectories}
\end{figure}

\begin{figure}[h]
    \centering
        \includegraphics[width=\textwidth]{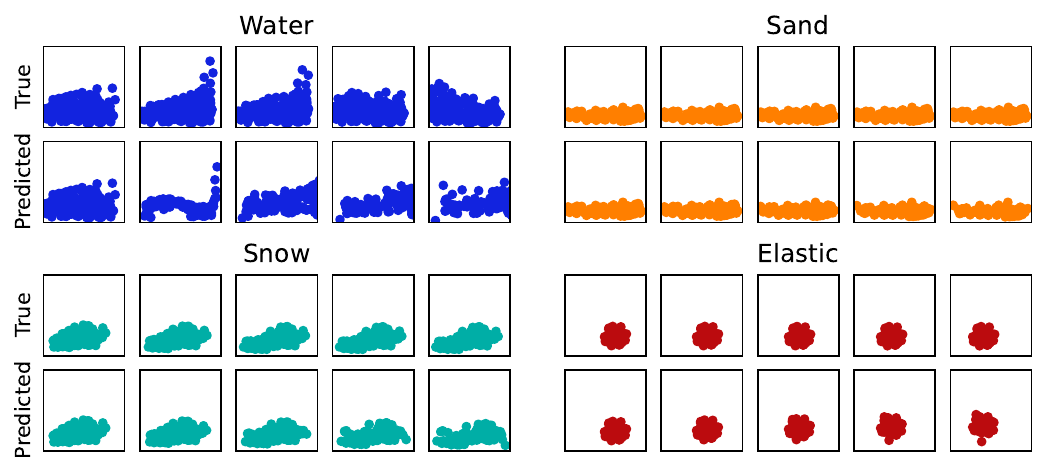}
    \caption{Comparison of true trajectories and trajectories predicted using our approach in the Fluid experiment. Time flows right.}
    \label{fig:fluid-video-encodings}
\end{figure}

Figure \ref{fig:dataset-trajectories} contains sample trajectories and their renderings, while the predicted trajectories are visualized in Figure \ref{fig:fluid-video-encodings}.

\section{Model Architecture}  \label{ap:model}
The \textsc{Video Encoder} processes a series of $64\times64$ RGB images. Each frame is encoded into a low-dimensional representation with a series of linear layers, each followed by a \verb|tanh| activation function. This output serves as the input for an LSTM with a 32-dimensional hidden state. Finally, the LSTM output is mapped to a 4-dimensional physical encoding.

Functions constituting a Graph Network are implemented as multi-layer perceptions with a single hidden layer and a \verb|tanh| activation. Table \ref{tab:fluid-model-summary} summarizes the model's architecture and exact dimensionality.

\begin{table*}
\caption{Construction and training parameters of the Video-Driven GNS for the Fluids task. $MLP(x;y;z)$ indicates a multi-layer perceptron with an $x$-dimensional input, $z$-dimensional output and hidden layers of size $y$.}
\label{tab:fluid-model-summary}
\centering
\SetTblrInner{rowsep=1pt,colsep=2pt}
\begin{tblr}{
  cells = {c},
  cell{2}{1} = {c=3}{},
  cell{6}{1} = {c=3}{},
  cell{9}{1} = {c=3}{},
  cell{13}{1} = {c=3}{},
  vline{2} = {1,3-6,7-9,10-13,13-15}{},
  vline{3} = {1,3-6,7-9,10-13,13-15}{},
  hline{2-16} = {-}{},
}
Parameter              & Value & Comments\\
\textbf{Video Encoder} &             & \\
Frame encoder          & $MLP(12,288;256;192;192)$ & \makecell{Flattening a 3-channel RGB frame\\of $64\times64$ resolution results\\in a 12,288-element vector} \\
\makecell{Hidden state\\of the LSTM}                       & 32    &       \\
Final mapping layer                      & $MLP(32,P=4)$    &       \\
\textbf{Graph Encoder} &          &   \\
Edge encoder                      & $MLP(10;48;48)$     &  \makecell{The initial vertex attribute consist\\of $C=3$ past two-dimensional velocities\\and 4-dimensional physical encoding $P$,\\resulting in a total of ten values.}  \\
Vertex encoder                 & $MLP(3;48;48)$         & \makecell{The direction of the edge requires\\two values. Involving the additional,\\scalar value, representing the length\\of the edge results\\in three values.} \\

\textbf{Graph Processor} & & \\
\makecell{Message passing\\steps ($M$)} & 3 & \\
Edge processor & $MLP(144,; 48; 48)$ & \makecell{144-dimensional input is created\\by concatenating two 48-dimensional\\edge embeddings and a single\\48-dimensional node embedding.} \\
Vertex processor & $MLP(96; 48; 48)$ & \makecell{96-dimensional input is created by\\concatenating\\a single 48-dimensional vertex embedding\\and an aggregation\\of 48-dimensional edge embeddings.}\\
\textbf{Graph Decoder} &  &  \\
Vertex decoder & $MLP(48; 48; 2)$ & \makecell{The output is a 2-dimensional\\vector representing acceleration.}

\end{tblr}
\end{table*}

% \section{Limitations} \label{ap:limitations}
% The \verb|sand| class has a parameter for the angle of friction, set to 45 degrees by default. When set to 0, the system behaves like the \verb|water| class. To test the model's ability to interpolate between systems, we create an additional evaluation dataset. This dataset includes 4 trajectories for 15 system classes, with the angle of friction increasing from 0 to 45 degrees. We then evaluate the system's one-step accuracy on this dataset. Figure \ref{fig:limitations-interpolation} shows that the model struggles with materials that were not present during training.
\end{document}